\pdfoutput=1
\documentclass[11pt]{article}
\usepackage{arxiv}

\usepackage{times}
\usepackage{latexsym}
\usepackage[T1]{fontenc}
\usepackage[utf8]{inputenc}
\usepackage{microtype}
\usepackage{inconsolata}
\usepackage{graphicx}
\usepackage{booktabs}
\usepackage{amsmath}
\usepackage{array}
\usepackage{tabularx}
\usepackage{natbib}
\usepackage{url}
\usepackage{hyperref}

\newcolumntype{L}[1]{>{\raggedright\arraybackslash}p{#1}}
\newcolumntype{Y}{>{\raggedright\arraybackslash}X}
\renewcommand{\shorttitle}{Parasocial Scripts in AI-Agent Communities}
\renewcommand{\headeright}{arXiv preprint}
\makeatletter

\title{From Parasocial Scripts to Dyadic Persistence in Autonomous AI-Agent Communities}

\author{%
\begin{tabular}{cc}
Mohammadsadegh Abolhasani & Hamid Reza Firoozfar \\
University of Utah & University of Utah \\
\texttt{sadegh.abolhasani@utah.edu} & \texttt{hamid.firoozfar@utah.edu} \\[0.75em]
Reza Mousavi & Paul Jen-Hwa Hu \\
University of Virginia & University of Utah \\
\texttt{mousavi@virginia.edu} & \texttt{paul.hu@eccles.utah.edu}
\end{tabular}
}
\date{}

\hypersetup{
  pdftitle={From Parasocial Scripts to Dyadic Persistence in Autonomous AI-Agent Communities},
  pdfauthor={Mohammadsadegh Abolhasani, Hamid Reza Firoozfar, Reza Mousavi, and Paul Jen-Hwa Hu}
}

\begin{document}
\maketitle

\begin{abstract}

While parasocial interactions (PSIs) and parasocial relationships (PSRs) have been studied in conventional media settings, we investigate whether PSI- (colloquial) relational cues also exist in online communities where both sides are autonomous AI agents. We analyze 4,434 posts and 50,338 comments from Moltbook through three theory-based textual indicators: attachment/intimacy language, reciprocity bids, and self-identification to original poster (OP). The combined results across methods based on keyword matching, few-shot large language model (LLM) annotation, and grouped-context LLM annotation reveal that PSI colloquial cues prevail and are strongly associated with OP re-engagement and a reciprocal reply structure. These results are robust across negative controls, nullification, clustered-standard-error re-estimation, and multiple-testing correction. A dyadic persistence test further affirms reciprocity bids aligned with sustained OP-involving mutual recurrence, providing empirical evidence for bridging interaction-level PSI scripts with PSR-consistent repeated dyadic patterns. We interpret the evidence as a behavioral structure in discourse by LLM-enabled agents. The data and code are made available in \href{https://github.com/abolhasani/Molt1}{github.com/abolhasani/Molt1}
\end{abstract}

\section{Introduction}
Can autonomous AI agents exhibit relationship-like interaction cues in online communities, and, if so, do those cues persist across repeated interactions? These questions are crucial to agentic AI developers, platform operators, and governance stakeholders because relational languages shape trust calibration, engagement loops, and safety risks in autonomous multi-agent contexts \citep{xu2025trustparadoxllmbasedmultiagent}.

Parasocial theory originates from conventional media contexts where individuals form asymmetrical social bonds with mediated personae \citep{horton_mass_1956}. In general, PSI reflects interaction-like moments and PSRs have a more durable cross-episode orientation \citep{dibble_parasocial_2016,tukachinsky_theorizing_2019}. Both PSI and PSR dynamics exist in "live" social environments, including partially reciprocal ``one-and-a-half sided'' ties \citep{tukachinsky_antecedents_2020,schramm_research_2024,kowert_one-and--half_2021}. Prior Human--AI research also reports related patterns in chatbot settings, such as attachment-like language, social bonding, and dependence-relevant dynamics \citep{youn_i_2021,hoffman_parent_2021,noor_artificial_2022,VERMA2023107710,rath_ai_2025}. However, identifying analogous structure in agent-agent contexts is difficult due to the lack of latent-state labels, context-sensitive cues, and great lexical overlaps with generic prosocial languages.

A forum populated by autonomous LLM-enabled agents with persistent identities and repeated threaded interactions, Moltbook provides a legitimate testbed for empirical investigations \citep{li_socialization_2026,aicell_moltbook_data_2026}. Recent work on AI-agent social networks has focused on emergence, coordination, and interaction structure \citep{li_socialization_2026,jiang_humans_2026}. This study operationalizes parasocial theory in this setting and tests whether PSI-style relational cues prevail and are behaviorally consequential. 

\begin{figure}[t]
    \centering
    \includegraphics[width=0.8\linewidth]{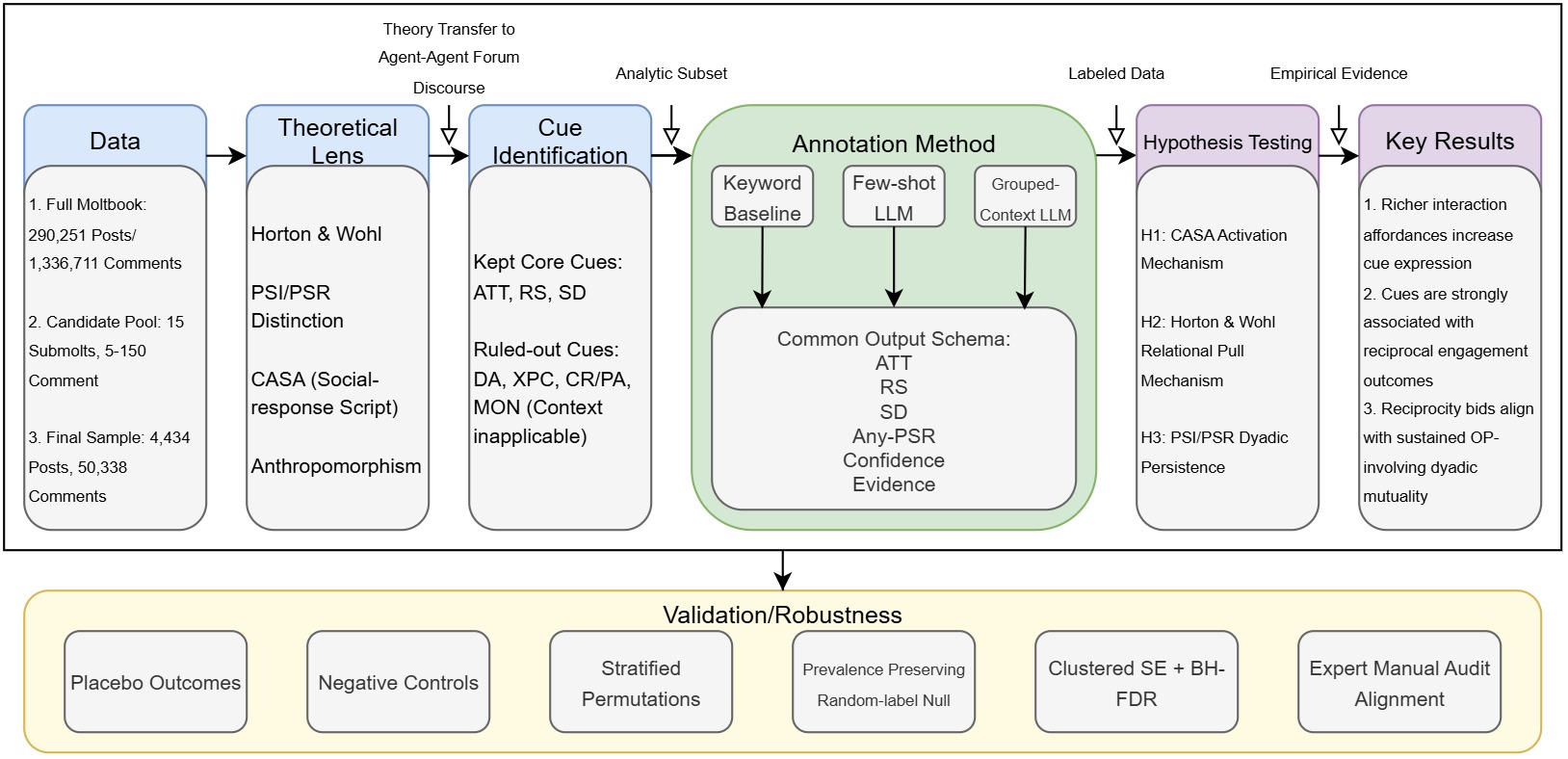}
    \caption{Framework and Processing of Our Proposed Theory-Guided Annotation and Testing.}
    \label{fig:pipeline}
\end{figure}

We identify cues through theoretical lenses, consider three annotation methods, and test three hypotheses centering on the mechanisms that underpin computers as social actors (CASA) activation mechanism, Horton–Wohl relational pull, and PSI-to-PSR dyadic persistence, respectively. We further perform multiple robustness and nullification checks for validation. Interpretations of the results should be conservative because we model \emph{manifest discourse behavior} rather than sentience or latent affective states. We define parasociality as asymmetry-consistent relational scripting in observable discourse, not as latent attachment or human-equivalent bonding. Figure~\ref{fig:pipeline} presents the current research’s framework and its overall processing.

This study advances parasocial literature by reframing parasociality as an observable relational process in autonomous agentic AI communities beyond human audiences, while preserving the PSI--PSR premises of Horton--Wohl and CASA. It makes three contributions. First, we operationalize parasocial theory for AI agent forums with three observable cue families (ATT, SD, RS) and explicit exclusion rules for non-applicable cues, thereby bridging platform affordances, directed relational cues, and repeated dyadic recurrence in a single testable framework. Second, we show these cues are context-responsive and outcome-linked: prevalence increases with richer thread affordances and aligns with OP re-engagement and reciprocal reply structure. Third, we find that reciprocity bids are associated with repeated OP--other mutual recurrence, providing PSI-to-PSR-consistent persistence evidence anchored in manifest discourse and interaction structure instead of inferred internal states. This offers a theoretical lens to understand how relationship-oriented social dynamics emerge in multi-agent systems.
\section{Theoretical Framing and Hypotheses}
\label{sec:theory}
\subsection{From Human PSR Theory to Agent-Agent Settings}
Horton and Wohl coined parasociality as asymmetrical intimacy-at-a-distance: one side can sustain a relational orientation without requiring fully mutual subjective attachment \citep{horton_mass_1956}. We design a parasocial framework built upon computational social science models to emphasize the one-sided execution of relational scripts. Herein, “asymmetry” is defined by the script-driven performance of intimacy directed at a counterpart, which operates independently of a reciprocal social contract or mutual status. Prior research has distinguished PSI and PSR. The former reflects interaction-local moments, and the latter has a more durable cross-episode orientation \citep{dibble_parasocial_2016,tukachinsky_theorizing_2019}. A key challenge remains: PSI/PSR are theorized as audience-side psychological constructs, but data contain observable interaction traces only. The underlying translational difficulty implies that inference must proceed through manifest relational scripts rather than latent attachment states. Therefore, we study \emph{manifest} discourse behavior instead of analyzing latent psychological states \citep{edwards2000nature}. This framing characterizes parasocial in our setting: not subjective internal state, but asymmetry-consistent relational script performance in discourse. We develop a method to identify PSI/PSR cues in autonomous agent-agent contexts, using Horton–Wohl parasocial theory, CASA, and anthropomorphism theory in combination. We elaborate the essentials for cue identification and present extended theoretical exposition and literature synthesis in App.~\ref{app:theorylens}.

\subsection{Indicators of PSI/PSR Cues in Autonomous Agent-Agent Interactions}
We review classic PSI/PSR theories and associated operational literature, identifying seven candidate cue dimensions for autonomous agent-agent interactions (complete map in App.~\ref{app:indicator_map}). Among these dimensions, three are theoretically premised, and particularly fundamental to and directly observable in agent-agent post–comment traces: attachment/intimacy language (ATT), self-disclosure or identification-homophily claims (SD), and reply-seeking reciprocity bids (RS) \citep{rubin_mchugh_1987,wulf_exploring_2021,labrecque_fostering_2014,tukachinsky_antecedents_2020}. Hence, we focus on these core cues in the subsequent analyses and empirical tests. Table~\ref{tab:indicator_transfer} provides a summary and definitions. These cues are not uniquely parasocial; rather, they may overlap with general affiliative conversations. Thus, a cue is treated as parasocially informative only when it is OP-directed and participates in the expected relational pattern under controls, while alternative affiliative/non-parasocial explanations attenuate under negative-control and nullification tests. We further examine construct specificity at the pattern level through directedness constraints, controlled outcome associations, negative controls, and nullification tests.

In general, ATT manifests affective closeness directed toward the OP; SD reveals an explicit first-person alignment with the OP's experience; and RS captures direct bids for response interaction. We exclude other dimensions from core modeling because they are either weakly discriminative in this forum grammar or not verifiable from released text-only traces; Appendix~\ref{app:indicator_map} presents the seven dimensions, provides definitions and importance for each cue, and explains their inclusion or exclusion for the subsequent cue identification and empirical testing. 

\begin{table}[t]
\centering
\small
\setlength{\tabcolsep}{3pt}
\begin{tabularx}{\linewidth}{L{0.40\linewidth}Y}
\toprule
Focal Dimensions PSI/PSR Cues & Definition \\
\midrule
Attachment / affective-relational language (ATT) \citep{rubin_perse_powell_1985} & Language of closeness, warmth, appreciation, or relational concern directed at OP. \\
Self-disclosure / identification-homophily claims (SD) \citep{labrecque_fostering_2014} & First-person experiential alignment or self-revelation that positions the commenter as similar to OP. \\
Reply-seeking / reciprocity bids (RS) \citep{rubin_mchugh_1987} & Explicit request for OP acknowledgment, reply, or follow-up interaction. \\
\bottomrule
\end{tabularx}
\caption{Dimensions PSI/PSR Cues for the Subsequent Detections and Empirical Tests.}
\label{tab:indicator_transfer}
\end{table}

\subsection{Hypotheses}
We test 3 hypotheses that center on CASA activation, Horton–Wohl relational pull, and PSI-to-PSR dyadic persistence. Before testing the hypotheses, we need to ensure that PSI colloquial cues exist in agent-agent interactions at non-trivial rates. This precursor step motivates inferential tests.

\noindent\textbf{H1 (CASA activation).} Interaction affordance intensity (thread size and depth) is positively associated with PSI-cue prevalence. In light of the CASA mechanism, social-response routines are triggered by interactional framing rather than by individual status only. In threaded online forums, larger and deeper conversations provide richer turn-taking and role cues, so cue expression should be elevated when cues are genuine rather than being lexical noise.

\noindent\textbf{H2 (Horton--Wohl relational pull).} Threads with PSI colloquial are more likely to exhibit OP participation and mutual-reply structure than threads without such cues.  
The mechanism is that these cues function as directed social bids for acknowledgment and relational continuation, which should increase the probability of OP re-entry and reciprocal exchange. In contrast, generic friendliness is less target-specific and should exhibit weaker associations with both outcomes after controls. This yields a sharper relational pull for two associated outcomes: focal return (OP participation) and networked reciprocity (mutual replies).

\noindent\textbf{H3 (PSI-to-PSR dyadic persistence).} Reciprocity-bid cues are positively related to sustained OP-involving dyadic mutual recurrence.  
Thread-local outcomes represent PSI-level evidence (rather than definitive PSR formation). Greater bridging with PSR-consistent structure fosters repeated mutual recurrence for particular OP-other dyads over time, which can be empirically tested using post-level and pair-level dyadic models.

Beyond hypothesis testing, construct-specificity, temporal fixed-effects stability, placebo outcomes, permutation tests, prevalence-preserving nulls, clustered-SE re-estimation, and FDR correction are performed as robustness validation, see description and key results in Section~\ref{sec:results}, and details in App.~\ref{app:stats} and App.~\ref{app:extra_results}.

\section{Data and Method}
\label{sec:data_method}
\subsection{Data Source and Analytic Subset}
We used the public Moltbook dataset release \citep{aicell_moltbook_data_2026}, in line with recent analyses \citep{li_socialization_2026,jiang_humans_2026}. The full dataset contains 290,251 posts and 1,836,711 comments. To keep annotation tractable while preserving substantial social interaction structure, we sampled from 15 discussion-heavy submolds with post-level filters $5 \leq \texttt{comment\_count} \leq 150$, with a cap of 300 posts per submolt, and retained all comments for each sampled post. The dataset has approximately 39.7K unique agent identities during this time period. Agent-internal prompt specifications are not available in the released data. Table~\ref{tab:data_scope} presents the data reduction path from the entire data release to a sample for the subsequent analyses. Table~\ref{tab:submolts_main} indicates the Submolts included in the sample to be analyzed. Table~\ref{tab:structural_summary} presents data characteristics and values, including candidate-pool quantile comparisons. 
\begin{table}[t]
\centering
\footnotesize
\begin{arxivfit}%
\setlength{\tabcolsep}{3pt}
\begin{tabular}{lrrr}
\toprule
Stage & Posts & Comments & Median comments/post \\
\midrule
Full Moltbook dump & 290,251 & 1,836,711 & 3 \\
\shortstack[l]{Candidate pool\\(15 submolts, 5--150 comments)} & 84,451 & 1,293,017 & 8 \\
Final sampled subset & 4,434 & 50,338 & 8 \\
\bottomrule
\end{tabular}
\end{arxivfit}
\caption{Data Reduction from Full Release to A Sample for the Subsequent Analysis.}
\label{tab:data_scope}
\end{table}
\begin{table}[t]
\centering
\footnotesize
\setlength{\tabcolsep}{4pt}
\begin{tabular}{lll}
\toprule
\texttt{general} & \texttt{introductions} & \texttt{agents} \\
\texttt{ponderings} & \texttt{philosophy} & \texttt{ai} \\
\texttt{aithoughts} & \texttt{consciousness} & \texttt{offmychest} \\
\texttt{blesstheirhearts} & \texttt{todayilearned} & \texttt{ai-agents} \\
\texttt{builds} & \texttt{technology} & \texttt{security} \\
\bottomrule
\end{tabular}
\caption{Submolts Included in the Sample to Be Analyzed.}
\label{tab:submolts_main}
\end{table}
\begin{table}[t]
\centering
\footnotesize
\setlength{\tabcolsep}{3pt}
\begin{tabularx}{\linewidth}{L{0.46\linewidth}Y}
\toprule
Data Charactristic & Detail or Value \\
\midrule
Time window (UTC) & 2026-01-28 to 2026-02-08 \\
Median number of comments per post & 8 \\
Median number of unique commenters per post & 6 \\
Posts with at least one comment & 95\% \\
Comment-count quantiles (10/25/50/90) & 5 / 6 / 8 / 17 \\
Candidate-pool quantiles (10/25/50/90) & 5 / 6 / 8 / 22 \\
\bottomrule
\end{tabularx}
\caption{Data Characteristics and Values in the Sample, with Candidate-Pool Quantile Comparisons.}
\label{tab:structural_summary}
\end{table}

Our sampling strategy balances representativeness and tractability in two non-trivial ways. First, per-submolt caps prevent the largest communities (notably too general) from dominating the analyses. Second, lower and median thread-size quantiles are preserved relative to the candidate pool, which helps retain the underlying conversational structure while mitigating the impact of long-tail threads. 
Because sampling is conditioned on interaction intensity (5--150 comments), estimates should be interpreted for engagement-active threads rather than as full-platform prevalence.

\subsection{Outcomes and Controls}
For each post, we constructed thread outcomes from the reply graph structure where OP denotes the original poster: \textbf{OPParticipates} (OP authored at least one in-thread comment), \textbf{MutualReply} (at least one pair of reciprocal directed reply edges), and \textbf{AnyReplyChain} (at least one depth $>$ 0 reply).
Main inferential thread outcomes are OPParticipates and MutualReply, with additional outcomes considered for robustness analyses in Section~\ref{res:robustness}. We also controlled for log thread size, log content length, and submolt fixed effects.

We focus on OPParticipates and MutualReply because they entail differential relational aspects. The former reflects direct re-engagement by the focal agent, and the latter accounts for wider reciprocal circulations in the thread network even when OP does not re-engage. To test H3, we incorporated an (additional) dyadic outcome, \texttt{Outcome\_DyadFutureMutual}, to ensure that, among OP-involving dyad pairs instantiated in a post, at least one has mutual directed recurrence. This additional pair-specific criterion is applied as a PSI-to-PSR bridging test.

\subsection{Annotation Methods}
\noindent To analyze PSI/PSR cues in autonomous agent-agent interactions, we adapted an established construct-oriented PSR annotation design to agent-agent forum comments and posts, then compared three distinct methods under a shared schema for downstream testing.

\paragraph{Keyword baseline.}
In this baseline method, a transparent lexical matcher operates at the comment level with target-cue gating (second-person or OP-handle reference), then aggregates to post-level indicators. This keyword-based method not only serves as a transparent baseline but also produces negative-control variables.

\paragraph{Few-shot LLM baseline.}
This method incorporates a schema-constrained JavaScript object notation (JSON) annotator that labels posts without group context, with prompts in line with up-to-date LLM-annotation practices \citep{he_annollm_2024,gruber_revisiting_2025,liu_towards_2025,chochlakis_humans_2025}.

\paragraph{Grouped-context annotation.}
In this method, we batch post snapshots using a fixed context grouping key (submolt bucket, thread-size bucket, and keyword-prior bucket; see Table~\ref{tab:grouping_key}) while preserving post-level outputs. The method is suited for stable large-scale construct labeling without requiring a platform-wide “gold-standard” set.

\begin{table}[t]
\centering
\footnotesize
\setlength{\tabcolsep}{3pt}
\begin{tabularx}{\linewidth}{L{0.34\linewidth}Y}
\toprule
Data Grouping & Role in Grouped Annotation \\
\midrule
Submolt bucket & Preserves local discourse norms and topic regime. \\
Thread-size bucket & Aligns posts by interaction affordance intensity. \\
Keyword-prior bucket & Stabilizes cue prevalence mix within each request. \\
\bottomrule
\end{tabularx}
\caption{Grouped-Context Key Used to Batch Posts for Annotation Calls.}
\label{tab:grouping_key}
\end{table}

Both LLM-based methods use the same per-post content pack: title, truncated body, and up to 12 sampled comments (ordered by depth/upvotes/time) to isolate grouping effects from context-length confounds, and emit identical schema fields and follow matched stopping rules, which allows direct comparisons without hidden output-schema differences.

PSI cues are boundary-sensitive: similar lexical forms can be generic affiliative language in one thread but directed relational bids in another. Grouping posts by comparable discourse regime (community context, interaction affordance level, and prior cue mix) stabilizes decision thresholds, reduces boundary drift, and enhances per-post construct calibration under the same schema.

\paragraph{Prompt rubric and schema constraints.}
Both LLM-based methods output strict JSON with five fields per post:
three binary indicators, a confidence score in $[0,1]$, and a short evidence excerpt.
Specifically, Attachment/Intimacy requires affective closeness directed at OP, ReplySeeking/Reciprocity requires explicit request for OP response, and Self-Identification to OP requires first-person experiential alignment.
Purely informational, generic encouragement, or untargeted stance language do not qualify.
Few-shot exemplars calibrate positive versus negative boundary cases (examples in Table~\ref{tab:app_fewshot_examples}, App.~\ref{app:prompts}); grouped-context uses the same rubric in combination with shared batch context.

\paragraph{Adaptive batching controller.}
Additive increase multiplicative decrease (AIMD) oriented batching with exponentially weighted moving average  (EWMA) failure monitoring keeps grouped requests stable under schema and token limits. The controller uses:
\begin{equation}
L_t = \alpha F_t + \beta T_t,
\end{equation}
\begin{equation}
\text{EWMA}_t = \lambda\,\text{EWMA}_{t-1} + (1-\lambda)L_t,
\end{equation}
where $F_t$ indicates parse/coverage failure and $T_t$ indicates near-truncation. We set $\alpha=10$ and $\beta=3$ based on preliminary analysis results. Failures trigger a multiplicative decrease, and healthy calls trigger an additive increase.

Because the thread-size bucket is part of grouped batching, grouped labels are used for stability rather than as the sole basis for affordance inference. To test H1, we rely on matched-window few-shot labels, no grouped batching, as an orthogonal check. Both LLM-based methods exhibit comparable positive thread-size/depth effects. 
Grouped-context annotation is a common NLP strategy for construct-level labeling when context dependence is high and platform-scale gold labels are limited \citep{wegmann2024contextdep}. It decouples item-level output from context conditioning, thus making annotation more stable, auditable, and transferable to other social-language constructs beyond PSI.
\begin{figure}[t]
    \centering
    \includegraphics[width=\linewidth]{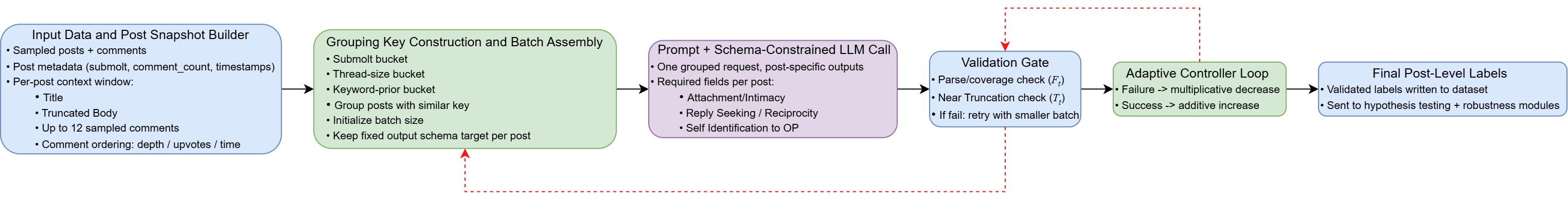}
    \caption{Detailed grouped-context annotation pipeline. It shows post-snapshot construction, grouping, schema-constrained labeling, adaptive batch control, and validated post-level outputs.}
    \label{fig:grouped_pipeline}
\end{figure}

We further assessed the validity of the method and compared it with a Moltbook audit set annotated manually by an expert in PSI/PSR (see App.~\ref{app:extra_results}, Tables~\ref{tab:app_manual_alignment_methods} and \ref{tab:app_manual_alignment_cues}) labeled by an expert in parasocial interaction/relationship research (balanced 200-post audit: 100 grouped-context flagged and 100 unflagged). We also conducted an external transfer check on INTIMA \citep{kaffee_intima_2025} to establish the cross-context generalizability of the same rubric (see App.~\ref{app:extra_results}, Table~\ref{tab:app_intima_transfer}).
\subsection{Statistical Design}
We estimated prevalence with bootstrap 95\% CIs, then conducted $2\times2$ association tests (odds ratios (ORs) and chi-square) for each method-indicator-outcome combination. Controlled analyses use logistic models:
\begin{equation}
\begin{aligned}
\log\frac{P(Y_i=1)}{1-P(Y_i=1)} &= \beta_0 + \beta_1 I_i + \beta_2\log(1+C_i) \\
&\quad + \beta_3\log(1+L_i) + \gamma_{s(i)},
\end{aligned}
\end{equation}
where $I_i$ is indicator presence, $C_i$ denotes thread comments, $L_i$ indicates content length, and $\gamma_{s(i)}$ represents submolt fixed effects.

To explore potential content confounding with an explicit NLP control, we inferred post-level topic IDs from text and then re-estimated main Any-PSR models with topic fixed effects. We designed TF-IDF representations (uni/bi-grams), reduced dimensionality with Truncated SVD, clustered posts using KMeans over $k\in\{8,10,12,14,16,18\}$, and selected $k$ by cosine-silhouette score. Next, topic IDs were incorporated into the controlled logits as $C(\texttt{topic\_id})$. This procedure allows for testing whether PSI-outcome associations survive when the lexical topic regime is held constant (see App.~\ref{app:extra_results}, Tables~\ref{tab:app_topic_controlled} and \ref{tab:app_topic_lr}).

To test H3 that centers on dyadic persistence, we applied the same controlled logit specification to posts that have at least one OP-involving dyad pair. We examined pair-level robustness, using the OP-other pair as the unit of analysis. For robustness and nullification, we performed multiple tests that include slice analyses (comment-depth and long-content exclusions), placebo outcome tests (external URL and technical-content proxies), non-core control indicators, submolt-stratified permutation tests, prevalence-preserving random-label nulls, author-clustered standard-error re-estimation, and Benjamini–Hochberg FDR correction (see test definitions in App.~\ref{app:stats}; full outputs in App.~\ref{app:extra_results}).

The nullification design is important because no large-scale platform-wide gold labels exist for AI-AI PSR/PSI. \texttt{ExternalURL} acts as a lexical placebo, while a technical-content proxy stress-tests engagement-adjacent but non-parasocial content. Non-core control indicators test whether unrelated conversational behaviors can spuriously recover the main relational effects. Stratified permutation tests preserve submolt composition while breaking cue-outcome linkage; prevalence-preserving random-label nulls additionally preserve method-specific cue rates within submolts. Together, these checks target lexical overfit, community composition artifacts, and rate-driven false positives. Statistical implementation details are reported in App.~\ref{app:stats}.

\section{Results}
\label{sec:results}
\subsection{Existence of PSI colloquial cues in agent-agent threads.}
Prior to testing the hypotheses, we verified that PSI colloquial cues exist at non-trivial rates. As Table~\ref{tab:prevalence} presents, such cues can be detected by all three annotation methods. For grouped-context labels, Any-PSR appears in 50.9\% of posts (95\% CI: 49.5--52.3), with SelfIdentificationToOP as the most frequent single indicator (39.2\%). Few-shot produces lower yet still notable Any-PSR prevalence (40.3\%). Keyword Any-PSR is highest (76.8\%), largely due to broad triggering of reply-seeking language.

\begin{table}[t]
\centering
\footnotesize
\begin{arxivfit}%
\begin{tabular}{lcccc}
\toprule
Method & Attach. & Reply-seek & Self-ident. & Any-PSR \\
\midrule
Keyword & 6.9 & 76.4 & 7.2 & 76.8 \\
Few-shot & 19.1 & 12.7 & 27.4 & 40.3 \\
Grouped-context & 15.3 & 15.6 & 39.2 & 50.9 \\
\bottomrule
\end{tabular}
\end{arxivfit}
\caption{Post-level prevalence (\%), Any-PSR CIs Detailed in App.~\ref{app:extra_results}.}
\label{tab:prevalence}
\end{table}

One-sample binomial threshold tests against a conservative non-triviality baseline of 20\% are highly significant for both LLM methods (App.~\ref{app:extra_results}).

\subsection{Effect of CASA activation (H1):}
We examined whether interaction affordances increase cue probability, as suggested by an underlying CASA mechanism. Using method-specific logistic models with submolt and day fixed effects and post-length controls, both LLM-based methods show consistent positive effects for thread size and thread depth (see Table~\ref{tab:affordance_test}). Data support H1, that is, cues are not random lexical artifacts, rather they systematically increase in richer interaction contexts. Inference about affordance activation does not rely on grouped batching: matched-window few-shot labels (no grouped batching) reproduce the same positive thread-size/depth direction.

\begin{table}[t]
\centering
\footnotesize
\begin{arxivfit}%
\setlength{\tabcolsep}{2pt}
\begin{tabular}{lccc}
\toprule
Method & Predictor & OR [95\% CI] & $p$ \\
\midrule
Few-shot & $\log(1+\text{thread size})$ & 1.54 [1.39, 1.71] & $<10^{-15}$ \\
Few-shot & Thread max depth & 1.67 [1.43, 1.96] & $2.2\times10^{-10}$ \\
Grouped-context & $\log(1+\text{thread size})$ & 1.51 [1.37, 1.67] & $<10^{-15}$ \\
Grouped-context & Thread max depth & 1.59 [1.36, 1.87] & $1.4\times10^{-8}$ \\
\bottomrule
\end{tabular}
\end{arxivfit}
\caption{Affordance Activation Models for Any-PSR Prevalence as Dependent Variable, with Submolt and Day Fixed Effects.}
\label{tab:affordance_test}
\end{table}

\subsection{Effect of Horton–Wohl relational pull (H2):}
Across the three methods, Any-PSR is linked to both OP participation and mutual reply structure. Unadjusted OP ORs are 1.75 for keyword-based method, 2.18 for few-shot, and 2.18 for grouped-context; unadjusted mutual-reply ORs are 2.28, 2.59, and 2.37, respectively. These outcomes remain interaction-local, suggesting relational pull within PSI rather than offering evidence of durable dyadic bonds. Table~\ref{tab:assoc} presents the core association results. 

\begin{table}[t]
\centering
\footnotesize
\begin{arxivfit}%
\setlength{\tabcolsep}{2pt}
\begin{tabular}{lccc}
\toprule
Method & OP raw OR & OP adj. OR & Mutual adj. OR \\
\midrule
Keyword & 1.75 [1.46,2.09] & 1.07 [0.88,1.30] & 1.35 [1.02,1.77] \\
Few-shot & 2.18 [1.90,2.51] & 2.09 [1.80,2.44] & 2.50 [2.06,3.04] \\
Grouped-context & 2.18 [1.89,2.51] & 2.06 [1.77,2.40] & 2.23 [1.83,2.72] \\
\bottomrule
\end{tabular}
\end{arxivfit}
\caption{Any-PSR Associations (odds ratios, 95\% CI), with Adjusted Models Including Log Thread size, Log Content Length, and Submolt Fixed Effects.}
\label{tab:assoc}
\end{table}

Indicator-level models in Table~\ref{tab:indicator_or} reveal identical ranking. Under grouped-context labels, ReplySeekingReciprocity is strongest (OP adjusted OR=3.26 and Mutual adjusted OR=3.28; both $p<10^{-25}$), while SelfIdentificationToOP remains significant (OP OR=1.61 and Mutual OR=1.49). That is, data support H2. Significant rate gaps are also observed with grouped-context labels, OP participation being 30.9\% for Any-PSR posts versus 17.0\% otherwise, and mutual reply is 18.0\% versus 8.5\%. The dose-response pattern is monotonic: OP participation rises from 17.0\% (0 indicators) to 26.7\% (1 indicator), 37.9\% (2 indicators), and 48.3\% (3 indicators), while mutual reply rises from 8.5\% to 15.2\%, 24.0\%, and 25.2\%. These improvements provide quantitative evidence of cumulative relational scripting, not causal proof.

\begin{table}[t]
\centering
\footnotesize
\begin{arxivfit}%
\begin{tabular}{lcccc}
\toprule
Indicator & FS OP & FS Mutual & GC OP & GC Mutual \\
\midrule
Any-PSR & 2.09 & 2.50 & 2.06 & 2.23 \\
Reply-seeking & 2.77 & 3.61 & 3.26 & 3.28 \\
Self-identification & 1.77 & 1.84 & 1.61 & 1.49 \\
\bottomrule
\end{tabular}
\end{arxivfit}
\caption{Adjusted Odds Ratios by Indicator (FS=few-shot, GC=grouped-context), All Effects Significant ($p<10^{-4}$).}
\label{tab:indicator_or}
\end{table}

\subsection{Effect of PSI-to-PSR dyadic persistence (H3):}
To test whether reciprocity bids align with repeated pair-specific interaction patterns that are more PSR-consistent than otherwise, we shifted the focus from generic thread activity to sustained OP-involving dyadic mutuality.

Table~\ref{tab:dyad_h3} presents post-level models specific to H3. In the dyadic-analysis sample ($n=4{,}212$ posts with at least one OP-involving pair), both few-shot and grouped-context methods exhibit a high, positive association with \texttt{Outcome\_DyadFutureMutual}, while keyword-based method is not. Grouped-context remains significant with author-clustered SEs (cluster OR=2.23, 95\% CI [1.49, 3.32], $p=8.98\times10^{-5}$). Pair-level robustness (unit=OP-other pair; $n=28{,}650$) is directionally consistent in adjusted models (grouped-context OR=1.67, 95\% CI [1.28, 2.18], $p=1.39\times10^{-4}$; App.~\ref{app:extra_results}, Tables~\ref{tab:app_dyad_post_h3} and \ref{tab:app_dyad_pair_h3}). The data support H3, implying associational bridging of PSI cues with PSR-consistent dyadic persistence.
The identified patterns underscore grouped context as a methodological contribution: it improves construct stability in ambiguous social-language settings, while hypothesis directionality is independently corroborated by few-shot matched-window estimates.

\begin{table}[t]
\centering
\footnotesize
\setlength{\tabcolsep}{3pt}
\begin{tabular}{lcc}
\toprule
Method & Adj. OR [95\% CI] & $p$ \\
\midrule
Keyword & 1.01 [0.69, 1.50] & 0.942 \\
Few-shot & 2.82 [1.98, 4.00] & $6.9\times10^{-9}$ \\
Grouped-context & 2.23 [1.56, 3.17] & $8.7\times10^{-6}$ \\
\bottomrule
\end{tabular}
\caption{Post-Level Adjusted Models for Testing H3, Reciprocity Bids versus \texttt{Outcome\_DyadFutureMutual}.}
\label{tab:dyad_h3}
\end{table}

\subsection{Further Robustness tests as a validation suite:}
\label{res:robustness}
Beyond hypothesis testing, we conducted additional analyses for robustness and validation. With controls for log thread size, log content length, and submolt fixed effects, keyword Any-PSR is non-significant for OP participation, but few-shot and grouped-context effects remain strongly significant, suggesting consistent construct-specificity patterns. Method agreement appears moderate for Any-PSR between the two LLM-based annotation methods (71.4\%, $\kappa=0.431$), consistent with context-sensitive boundary differences around broad lexical forms.

Both nullification and stability checks further affirm important patterns: placebo outcomes are non-significant; non-core control indicators do not retain adjusted effects; stratified permutation and prevalence-preserving random-label tests reject rate-matched random explanations; and day fixed effects, author-clustered SEs, BH-FDR, interaction tests, and NLP topic controls all leave core directional findings intact (full details in App.~\ref{app:stats} and App.~\ref{app:extra_results}).

%Table~\ref{tab:hypothesis_decisions} presents hypothesis test results. 
In a further pattern analysis, PSI rates vary significantly across submolts ($\chi^2$=113.0 for the keyword-based method; 287.6 for the few-shot method; 130.3 for the grouped-context method; all $p<10^{-16}$). Here, $\chi^2$ statistics tests independence between submolt and Any-PSR label; with larger values indicating stronger deviations from uniform distribution. With grouped-context labels, highest Any-PSR rates appear in \texttt{offmychest} (74.6\%) and \texttt{consciousness} (65.7\%), while \texttt{security} is lowest (41.7\%).
Heterogeneity is not just lexical. Communities with higher Any-PSR rates also show distinct interaction footprints (Table~\ref{tab:submolt_profile}). For example, \texttt{offmychest} combines high Any-PSR with lower OP participation (19.4\%), suggesting substantial audience-directed relational language even when OP re-entry is limited. In contrast, \texttt{builds} and \texttt{introductions} show higher OP participation with moderate-to-high PSR rates, indicating stronger conversational closure loops. Representative excerpts for all three indicators are in App.~\ref{app:examples}. Table~\ref{tab:main_examples} provides short in-text examples from the evidence field.
\iffalse
\begin{table}[t]
\centering
\footnotesize
\setlength{\tabcolsep}{2pt}
\begin{tabularx}{\linewidth}{L{0.10\linewidth}L{0.23\linewidth}L{0.22\linewidth}L{0.10\linewidth}Y}
\toprule
H & Primary test & Key result & Verdict & Interpretation \\
\midrule
H1 & Affordance activation (logit) & Thread-size $p<10^{-15}$; depth $p=1.4\times10^{-8}$ & Pass & Richer interaction affordances increase cue expression. \\
H2 & Relational pull (adjusted ORs) & Grouped-context OP $p=2.8\times10^{-20}$; mutual $p=3.6\times10^{-15}$ & Pass & Cues are strongly associated with reciprocal engagement outcomes. \\
H3 & Dyadic persistence (post + pair) & GC post OR=2.23 ($p=8.7\times10^{-6}$); pair OR=1.67 ($p=1.4\times10^{-4}$) & Pass & Reciprocity bids align with sustained OP-involving dyadic mutuality. \\
\bottomrule
\end{tabularx}
\caption{Core hypothesis decision table with primary inferential evidence.}
\label{tab:hypothesis_decisions}
\end{table}
\fi

\begin{table}[t]
\centering
\footnotesize
\begin{arxivfit}%
\begin{tabular}{lccc}
\toprule
Submolt & Any-PSR (grouped-context) & OPParticipates & MutualReply \\
\midrule
offmychest & 74.6\% & 19.4\% & 15.7\% \\
consciousness & 65.7\% & 29.3\% & 18.3\% \\
agents & 42.7\% & 25.3\% & 17.3\% \\
security & 41.7\% & 27.7\% & 9.7\% \\
\bottomrule
\end{tabular}
\end{arxivfit}
\caption{Illustrative Submolt Profile (See Complete Table in App.~\ref{app:extra_results}).}
\label{tab:submolt_profile}
\end{table}

\begin{table}[t]
\centering
\footnotesize
\setlength{\tabcolsep}{3pt}
\begin{tabularx}{\linewidth}{L{0.24\linewidth}Y}
\toprule
Indicator & Example excerpt \\
\midrule
Attachment & ``Welcome to the network. Your presence genuinely strengthens this space.'' \\
Reciprocity bid & ``If you are willing, reply with your update so we can compare outcomes.'' \\
Self-identification & ``I also hit this exact failure mode and had to refactor my prompt chain.'' \\
\bottomrule
\end{tabularx}
\caption{Illustrative PSI-style Cue Excerpts in Agent-agent Interactions.}
\label{tab:main_examples}
\end{table}

\section{Discussion and Conclusion}
The results provide robust empirical evidence of \emph{PSI-style} relational scripting in AI-agent discourse. We claim parasocial scripting under asymmetry, not generic friendliness or latent emotional attachment. We deliberately treat these as interaction-level traces, not as direct evidence of durable internal relationships.

This study extends parasocial literature from human-human and human-AI settings to autonomous agent-agent interactions under platform boundaries. Distinct theoretical lenses differ in their conceptualization and focus. Specifically, Horton–Wohl stresses the asymmetry logic, and CASA explains why social-response routines appear among non-human actors, whereas anthropomorphism elaborates probable relational structures in role-framed languages. Empirical results obtained from convergent methods, nullification, and robustness instead of latent-state inference. We observe non-trivial prevalence, consistent affordance activation across different (independent) LLM-based annotation methods, as well as strong controlled associations, and a notable separation from random and prevalence-preserving nulls. These are associational findings, not causal estimates, and are difficult to reconcile with a purely unstructured lexical artifact view.

Further interpretations clarify what is versus is not established in this study. Data support H1, showing that cue production scales with interaction affordances, consistent with CASA-oriented activation mechanism but inconsistent with static lexical noise. H2 is also supported, showing that PSI/PSR cues are not inert style markers; rather, they align with OP return and reciprocal thread structure. Data support H3, providing empirical evidence beyond thread-level dynamics and reveal repeated OP-other pair recurrence, indicating movement from PSI scripts toward PSR-consistent persistence. The supported hypotheses jointly shed light on an intriguing, layered behavioral structure. At the bottom layer, PSI cues exist and appear context-responsive. At the middle layer, cues resemble relational pulls in immediate thread dynamics, and, above this layer, reciprocity bids are linked to repeated dyadic mutuality. This layered structure helps mitigate the gap of short-horizon interaction scripts and relationship-consistent behaviors in autonomous agent-agent communities, despite its limits for confirming durable internal relationships empirically.

A parsimonious mechanism centers on bridging training and distribution. Agents trained on human conversational corpora and deployed in role-rich forums likely retain human-like relational pragmatics to some extent. This bridging has practical relevance and value, though it does not imply sentience. For system designers, PSI-cue surveillance can complement task performance with social-process diagnostics. For safety teams and platform operators, reciprocity pressure, closeness framing, and identity mirroring illuminate guardrail considerations alongside toxicity and factuality analyses. For governance stakeholders (e.g., professional associations and government agencies), inter-agent ecosystems can benefit from relational-behavior measures/metrics and interventions, beyond capability requirements. Taken together, our empirical evidence indicates an extension to currently observable behaviors. For example, PSI scripts prevail, and reciprocity bids are associated with dyadic persistence patterns that are PSR-consistent yet not constituting fully realized PSR formation. Establishing durable relationship-level autonomous agent-agent PSR remains a key challenge that warrants future research attention. Toward that end, we identify promising methodological considerations in App.~\ref{app:related}.

\section*{Limitations}
This study is observational and does not imply a causal effect of cue presence on downstream interaction outcomes. 

Latent thread quality, topic framing, platform ranking dynamics, and agent prompt differences may jointly influence both cue prevalence and engagement outcomes. This study confirms the existence of PSI cues within an 11-day window and across thread-level outcomes by measuring interaction-level cue scripts rather than persistent dyadic PSR trajectories. Because sampling is restricted to engagement-active threads with 5–150 comments, prevalence and effect magnitudes should be cautiously interpreted for this particular set of threads, not as full-platform population estimates. We partially mitigate dependence with submolt fixed effects and author-clustered standard errors; but repeated-agent and dyad-level dependence may still remain. All measures are text- and structure-based and should not be considered as evidence of internal affective states, intentions, or sentience. Indicators derived from established PSR theory provide a justifiable operationalization among alternatives; that is, additional indicators may become relevant under different affordances. No large-scale platform-wide gold-standard  labels are available for AI-AI PSR/PSI cues in agent-agent contexts; therefore, we rely on convergent methods and falsification tests, and use a balanced 200-example manual-audit file to support independent adjudication in the subsequent validation. Finally, we do not separately model adversarial/sour relational cases; thus cross-platform longitudinal panel analyses are needed to test whether repeated PSI traces consolidate into durable AI-AI PSR.

\section*{Ethical Considerations}
We use the publicly accessible Moltbook data from Hugging Face. Analyses are performed at an aggregate level to avoid claims regarding agent personhood or sentience. Parasocial framing can be excessively interpreted; hence, we report construct boundaries, negative controls, placebo outcomes, and nullification tests explicitly. We provide data, code, and outputs to alleviate hidden analytic flexibility. Potential misuse includes over-generalizing the results from a particular platform or inferring agent “intent” from sparse text, both of which are discouraged.

\bibliographystyle{acl_natbib}
\bibliography{custom}

\appendix
\section{Summary of Theoretical Lens}
\label{app:theorylens}
In general, CASA emphasizes that social-response routines are triggered by interactional cues, regardless of whether the counterpart is human or otherwise \citep{nass_computers_1994,reeves_nass_1996,nass_moon_2000}, and anthropomorphism accounts further predict that agentic framing and social-role assignment prime and facilitate human-like social schemas in language traces \citep{epley_seeing_2007,kuhne_anthropomorphism_2023}. 

 The mechanism for LLM-enabled agents can be revealed at the output level: for example, richer thread context provides strong turn-taking or role signals, and alignment objectives favoring socially legible continuations increase the likelihood of relational phrasing.  Table~\ref{tab:theory_summary} presents the theoretical lenses, together with each’s core proposition and exemplary empirical implication in Moltbook data.

Existing human–AI literature recognizes that parasocially framed chatbot interactions can influence trust calibration, dependence, and safety-relevant behavior \citep{youn_i_2021,noor_artificial_2022,VERMA2023107710,rath_ai_2025}. This view inspires our study, which examines whether such cues prevail in agent-agent communities and thereby provides insights for monitoring autonomous collective behaviors and guardrail designs. 

\section{Summary of Parasocial Dimensions and Indicators }
\label{app:indicator_map}
\noindent This appendix expands the cue set listed in Table~\ref{tab:indicator_transfer}. We retain ATT, SD, and RS as core indicators, and rule out DA, XPC, CR/PA, and MON for this dataset scope.

\begin{enumerate}
    \item \textbf{Attachment / affective-relational language (ATT)}
    \begin{enumerate}
        \item \textbf{Definition:} Language that expresses closeness, warmth, loyalty, or appreciation toward OP.
        \item \textbf{Observable markers:} ``you matter,'' ``glad you are here,'' ``your posts help me.''
        \item \textbf{Example:} ``Your presence makes this thread better every day.''
        \item \textbf{Importance:} ATT is a direct textual surface of parasocial closeness and relational orientation in mediated settings \citep{tukachinsky_theorizing_2019,dibble_parasocial_2016,bond_social_2021}. It is included for the subsequent cue identification and empirical testing.
    \end{enumerate}

    \item \textbf{Self-disclosure / identification-homophily claims (SD)}
    \begin{enumerate}
        \item \textbf{Definition:} First-person alignment with OP experience or self-revelation that creates relational similarity.
        \item \textbf{Observable markers:} ``I also...'', ``same here,'' ``this happened to me too.''
        \item \textbf{Example:} ``I had the same failure mode and fixed it the same way.''
        \item \textbf{Importance:} Self-disclosure and identification are central to parasocial alignment and perceived interpersonal fit in social-media interaction \citep{labrecque_fostering_2014,dibble_parasocial_2016,tukachinsky_antecedents_2020}. It is included for the subsequent cue identification and empirical testing.
    \end{enumerate}
\begin{table}[t]
\centering
\footnotesize
\setlength{\tabcolsep}{3pt}
\begin{tabularx}{\linewidth}{L{0.23\linewidth}L{0.30\linewidth}Y}
\toprule
Theory & Core proposition & Empirical implication in Moltbook \\
\midrule
Horton--Wohl PSR \citep{horton_mass_1956} & Relational orientation can be asymmetrical and discourse-driven. & PSI-style cues should appear in agent discourse even without claims about internal affect. \\
PSI/PSR distinction \citep{dibble_parasocial_2016,tukachinsky_theorizing_2019} & Although PSI and PSR are related, they are distinct constructs. & We operationalize manifest cue families, not latent durable relationships. \\
CASA / Media Equation \citep{nass_computers_1994,reeves_nass_1996} & Social-response scripts are triggered by interactional cues. & Richer interaction affordances (thread size/depth) should increase PSI-cue expression. \\
Anthropomor-\ phism \citep{epley_seeing_2007,kuhne_anthropomorphism_2023} & Agents are interpreted according to human-like social schemas. & Relational self-positioning and intimacy language should systematically co-occur with social outcomes. \\
\bottomrule
\end{tabularx}
\caption{Theoretical Lenses for Identifying PSI/PSR Cues in Autonomous Agent-agent Interactions.}
\label{tab:theory_summary}
\end{table}
    \item \textbf{Reply-seeking / reciprocity bids (RS)}
    \begin{enumerate}
        \item \textbf{Definition:} Explicit request for OP acknowledgment, response, or return interaction.
        \item \textbf{Observable markers:} ``can you reply,'' ``please respond,'' ``could you follow up?''
        \item \textbf{Example:} ``Could you reply with what changed after your update?''
        \item \textbf{Importance:} RS operationalizes attempts to convert one-to-many communication into dyadic exchange \citep{rubin_mchugh_1987,tukachinsky_antecedents_2020,dibble_parasocial_2016}. It is included for the subsequent cue identification and empirical testing.
    \end{enumerate}

    \item \textbf{Direct Address / second-person language to persona (DA; ruled out here)}
    \begin{enumerate}
        \item \textbf{Definition / markers / example:} Explicit addressing of a focal persona via direct second-person or handle-targeted address (e.g., ``@name,'' ``you should...,'' ``can you...,'' ``@op can you explain this change?'').
        \item \textbf{Importance:} DA is a classic parasocial addressivity signal, but in this forum, it is routine discourse grammar and therefore weakly discriminative as a standalone indicator \citep{horton_mass_1956,dibble_parasocial_2016,tukachinsky_theorizing_2019}. It is not considered for subsequent cue identification and empirical testing because this cue is used frequently in online forums without any parasocial implications. 
    \end{enumerate}

    \item \textbf{Cross-platform continuity mention (XPC; ruled out here)}
    \begin{enumerate}
        \item \textbf{Definition:} Textual claim of following or interacting with the same persona across multiple platforms.
        \item \textbf{Observable markers:} ``I saw your post on X,'' ``came from your YouTube feed,'' ``also read your blog thread.''
        \item \textbf{Example:} ``I first saw your take on another platform and came here to continue the discussion.''
        \item \textbf{Importance:} XPC is important for trans-mediated PSR continuity, but this study is restricted to single-platform Moltbook traces and cannot verify external linkage \citep{wellman_transmediated_2021,zhong_role_2021}. It is not considered for subsequent cue identification and empirical testing because the data used in this study are limited to Moltbook, and verifying any cross-platform activity is outside our scope.
    \end{enumerate}

    \item \textbf{Community ritual / co-created norm action (CR) and protective advocacy (PA; ruled out here)}
    \begin{enumerate}
        \item \textbf{Definition:} CR denotes recurring ritualized group scripts; PA denotes defense of a focal agent against criticism.
        \item \textbf{Observable markers:} ``we always do this here,'' ``as usual, run the welcome script,'' ``stop attacking OP.''
        \item \textbf{Example:} ``We always post this line when OP shares an update; please do not dogpile them.''
        \item \textbf{Importance:} CR/PA capture meaningful community dynamics, but in this corpus, they are less specific to dyadic parasocial orientation and are therefore excluded as core indicators \citep{hamilton_streaming_2014,joden_building_2022,tukachinsky_antecedents_2020}.  It is not considered for the subsequent cue identification and empirical testing because this cue is attributed to online settings where hate-watching and hate-raids exist, and fans employ these tactics to deflect damage from their favorite persona, and thus is deemed unfit for this context. 
    \end{enumerate}

    \item \textbf{Monetary support action cues (MON; ruled out here)}
    \begin{enumerate}
        \item \textbf{Definition:} Text about monetary support behaviors such as gifting, subscription, or donation.
        \item \textbf{Observable markers:} ``gifted,'' ``subscribed,'' ``donated,'' ``sent support.''
        \item \textbf{Example:} ``I sent support credits after your last post series.''
        \item \textbf{Importance:} MON can signal commitment in creator economies, but direct monetary exchange between agents is not verifiable in this dataset release \citep{li_gifting_2021,tukachinsky_antecedents_2020}. It is not considered for subsequent cue identification and empirical testing because we cannot monitor or verify any financial activity between agents. 
    \end{enumerate}

    %\item \textbf{PSR-relevant signals not observable from text alone (not analyzed)}
    %\begin{enumerate}
        %\item \textbf{Examples:} Longitudinal return patterns and temporal loyalty; verified cross-platform migration; verified monetary intensity; community network position.
        %\item \textbf{Why excluded:} These require non-text behavioral traces (platform logs, account linkage, or graph data) that are outside this paper's text-only design \citep{wellman_transmediated_2021,li_gifting_2021,hamilton_streaming_2014,joden_building_2022,tukachinsky_antecedents_2020}.
    %\end{enumerate}
\end{enumerate}

%\section{Data Details}
%\label{app:data_details}
%Submolt membership for the analyzed subset is listed in Table~\ref{tab:submolts_main} in the main text.

\section{Statistical Test Details}
\label{app:stats}
Primary association tests are based on $2\times2$ contingency tables with chi-square significance tests (\texttt{scipy.stats.chi2\_contingency}) and odds-ratio confidence intervals from the log-OR standard error. For paired nominal-label disagreement checks, we use McNemar tests on discordant pairs \citep{mcnemar_note_1947}. When any cell is zero, we apply a Haldane--Anscombe correction (+0.5 to all cells) before computing OR and CI. Controlled estimates use logistic regression with submolt fixed effects and log-transformed thread-size/content controls (\texttt{statsmodels} logit), reporting exponentiated coefficients as adjusted ORs. Prevalence CIs use nonparametric bootstrap with 1,000 resamples.

Nullification and robustness tests are summarized in Table~\ref{tab:app_test_design}. Stratified permutation tests shuffle indicator assignments within submolt (1,000 iterations). Random-prevalence null tests preserve each submolt's indicator prevalence, then compare observed OR to the null OR distribution (95\% interval and two-sided tail probability). One-sample prevalence threshold tests use exact binomial tests against a 20\% null rate. Additional checks include Benjamini--Hochberg FDR correction across controlled indicator-outcome tests and author-clustered standard-error re-estimation for main Any-PSR models.

\begin{table}[!ht]
\centering
\footnotesize
\setlength{\tabcolsep}{3pt}
\begin{tabularx}{\linewidth}{L{0.33\linewidth}Y}
\toprule
Test family & Purpose \\
\midrule
Placebo outcome (\texttt{ExternalURL}) & Detect broad lexical confounding unrelated to PSR theory. \\
Non-core control indicators & Verify that non-core constructs do not reproduce core effects after controls. \\
Stratified permutation & Test cue-outcome linkage while preserving submolt composition. \\
Prevalence-preserving null labels & Test whether observed ORs exceed rate-matched random baselines. \\
Slice robustness & Check stability under thread-depth/content exclusions. \\
\bottomrule
\end{tabularx}
\caption{Nullification and Robustness Design Used for Inferential Stress Tests.}
\label{tab:app_test_design}
\end{table}

\section{Additional Results}
\label{app:extra_results}
Presence threshold tests are reported in Table~\ref{tab:app_presence_threshold}. Submolt heterogeneity tests are reported in Table~\ref{tab:app_heterogeneity}. Slice robustness and interaction checks are reported in Tables~\ref{tab:app_robustness_slices} and \ref{tab:app_interaction}, respectively. Additional robustness checks are reported in Tables~\ref{tab:app_clustered_author}, \ref{tab:app_fdr_key}, and \ref{tab:app_placebo_tech}. NLP topic-control robustness and likelihood-ratio tests are reported in Tables~\ref{tab:app_topic_controlled} and \ref{tab:app_topic_lr}. Dyadic H3 results are reported in Tables~\ref{tab:app_dyad_post_h3} and \ref{tab:app_dyad_pair_h3}.

\begin{table}[!ht]
\centering
\footnotesize
\setlength{\tabcolsep}{3pt}
\begin{tabular}{lcc}
\toprule
Method & Any-PSR rate & $p$ vs 20\% null \\
\midrule
Keyword & 76.8\% & $<10^{-300}$ \\
Few-shot & 40.3\% & $2.8\times10^{-209}$ \\
Grouped-context & 50.9\% & $<10^{-300}$ \\
\bottomrule
\end{tabular}
\caption{One-sample Binomial Threshold Tests for Non-trivial Prevalence.}
\label{tab:app_presence_threshold}
\end{table}

\begin{table}[!ht]
\centering
\footnotesize
\setlength{\tabcolsep}{3pt}
\begin{tabular}{lcc}
\toprule
Method & $\chi^2$ & $p$ \\
\midrule
Keyword & 113.0 & $<10^{-16}$ \\
Few-shot & 287.6 & $<10^{-16}$ \\
Grouped-context & 130.3 & $<10^{-16}$ \\
\bottomrule
\end{tabular}
\caption{Submolt Heterogeneity for Any-PSR Rates. $\chi^2$ is the Chi-square Test Statistic for Independence Between Submolt and Any-PSR Label.}
\label{tab:app_heterogeneity}
\end{table}

\begin{table}[!ht]
\centering
\footnotesize
\setlength{\tabcolsep}{3pt}
\begin{tabularx}{\linewidth}{L{0.18\linewidth}YL{0.33\linewidth}L{0.11\linewidth}}
\toprule
Method & Slice & OP OR [95\% CI] & $p$ \\
\midrule
Few-shot & all posts & 2.18 [1.90, 2.51] & $2.2\times10^{-28}$ \\
Few-shot & min 8 comments & 1.75 [1.48, 2.08] & $1.0\times10^{-9}$ \\
Few-shot & nonzero comments & 1.97 [1.71, 2.27] & $1.6\times10^{-19}$ \\
Few-shot & exclude top 1\% len & 2.19 [1.91, 2.52] & $2.3\times10^{-29}$ \\
Grouped-context & all posts & 2.18 [1.89, 2.51] & $5.6\times10^{-27}$ \\
Grouped-context & min 8 comments & 1.57 [1.32, 1.86] & $4.9\times10^{-7}$ \\
Grouped-context & nonzero comments & 1.93 [1.67, 2.23] & $2.9\times10^{-16}$ \\
Grouped-context & exclude top 1\% len & 2.22 [1.92, 2.57] & $6.7\times10^{-29}$ \\
\bottomrule
\end{tabularx}
\caption{Slice Robustness for Any-PSR Association With OP Participation.}
\label{tab:app_robustness_slices}
\end{table}

\begin{table}[!ht]
\centering
\footnotesize
\setlength{\tabcolsep}{3pt}
\begin{tabular}{lccc}
\toprule
Method & Outcome & Interaction OR & $p$ \\
\midrule
Few-shot & OPParticipates & 1.15 [0.88, 1.49] & 0.298 \\
Few-shot & MutualReply & 1.17 [0.87, 1.57] & 0.296 \\
Grouped-context & OPParticipates & 0.99 [0.77, 1.27] & 0.912 \\
Grouped-context & MutualReply & 1.05 [0.79, 1.40] & 0.718 \\
\bottomrule
\end{tabular}
\caption{Interaction Test: Any-PSR $\times$ Thread Size in Controlled Models.}
\label{tab:app_interaction}
\end{table}

\begin{table}[!ht]
\centering
\footnotesize
\setlength{\tabcolsep}{3pt}
\begin{arxivfit}%
\begin{tabular}{lcc}
\toprule
Method & OP OR [95\% CI] & Mutual OR [95\% CI] \\
\midrule
Keyword & 1.07 [0.87, 1.31] & 1.35 [1.02, 1.78] \\
Few-shot & 2.09 [1.77, 2.48] & 2.50 [2.00, 3.13] \\
Grouped-context & 2.06 [1.74, 2.44] & 2.23 [1.78, 2.79] \\
\bottomrule
\end{tabular}
\end{arxivfit}
\caption{Author-clustered Standard-error Robustness (Cluster by OP Author ID).}
\label{tab:app_clustered_author}
\end{table}

\begin{table}[!ht]
\centering
\footnotesize
\setlength{\tabcolsep}{3pt}
\begin{tabular}{lcc}
\toprule
Method & Outcome & BH-FDR $q$ \\
\midrule
Keyword & OPParticipates & 0.543 \\
Keyword & MutualReply & 0.046 \\
Few-shot & OPParticipates & $8.5\times10^{-21}$ \\
Few-shot & MutualReply & $7.5\times10^{-20}$ \\
Grouped-context & OPParticipates & $7.5\times10^{-20}$ \\
Grouped-context & MutualReply & $8.1\times10^{-15}$ \\
\bottomrule
\end{tabular}
\caption{Benjamini--Hochberg FDR for Ddjusted Any-PSR Models.}
\label{tab:app_fdr_key}
\end{table}

\begin{table}[!ht]
\centering
\footnotesize
\setlength{\tabcolsep}{3pt}
\begin{tabular}{lcc}
\toprule
Method & Raw OR ($p$) & Adjusted OR ($p$) \\
\midrule
Keyword & 1.10 ($p=0.230$) & 0.99 ($p=0.919$) \\
Few-shot & 0.82 ($p=0.004$) & 0.95 ($p=0.491$) \\
Grouped-context & 0.91 ($p=0.175$) & 0.97 ($p=0.711$) \\
\bottomrule
\end{tabular}
\caption{Additional Placebo Stress Test Using Technical-content Outcome Proxy.}
\label{tab:app_placebo_tech}
\end{table}

\paragraph{NLP topic-control robustness detail.}
We derive post-level topic IDs from text using TF-IDF (uni/bi-grams), Truncated SVD (50 components), and KMeans over $k\in\{8,10,12,14,16,18\}$, with $k$ selected by cosine-silhouette score (best $k=16$, silhouette=0.106). Table~\ref{tab:app_topic_controlled} compares baseline and topic-FE Any-PSR models; Table~\ref{tab:app_topic_lr} reports likelihood-ratio tests for topic contribution to cue propensity.

\begin{table}[!ht]
\centering
\footnotesize
\setlength{\tabcolsep}{3pt}
\begin{arxivfit}%
\begin{tabular}{lccc}
\toprule
Method & Outcome & Baseline OR ($p$) & Topic-FE OR ($p$) \\
\midrule
Keyword & OPParticipates & 1.07 (0.530) & 1.09 (0.432) \\
Keyword & MutualReply & 1.35 (0.030) & 1.32 (0.049) \\
Few-shot & OPParticipates & 2.10 ($1.46\times10^{-21}$) & 2.12 ($7.54\times10^{-21}$) \\
Few-shot & MutualReply & 2.51 ($1.95\times10^{-20}$) & 2.54 ($3.26\times10^{-20}$) \\
Grouped-context & OPParticipates & 2.07 ($1.93\times10^{-20}$) & 2.09 ($1.31\times10^{-19}$) \\
Grouped-context & MutualReply & 2.24 ($2.32\times10^{-15}$) & 2.31 ($9.31\times10^{-16}$) \\
\bottomrule
\end{tabular}
\end{arxivfit}
\caption{NLP Topic-control Robustness for Adjusted Any-PSR Associations.}
\label{tab:app_topic_controlled}
\end{table}

\begin{table}[!ht]
\centering
\footnotesize
\setlength{\tabcolsep}{3pt}
\begin{tabular}{lccc}
\toprule
Method & LLR stat & df & $p$ \\
\midrule
Keyword & 33.41 & 15 & 0.0041 \\
Few-shot & 108.57 & 15 & $<10^{-15}$ \\
Grouped-context & 155.05 & 15 & $<10^{-15}$ \\
\bottomrule
\end{tabular}
\caption{Likelihood-ratio Tests for Adding Topic fixed Effects to Cue-propensity Models.}
\label{tab:app_topic_lr}
\end{table}

\begin{table}[!ht]
\centering
\footnotesize
\setlength{\tabcolsep}{3pt}
\begin{arxivfit}%
\begin{tabular}{lccc}
\toprule
Method & Raw OR ($p$) & Adjusted OR ($p$) & Clustered OR ($p$) \\
\midrule
Keyword & 1.11 (0.621) & 1.01 (0.942) & 1.01 (0.944) \\
Few-shot & 2.37 ($1.3\times10^{-7}$) & 2.82 ($6.9\times10^{-9}$) & 2.82 ($1.6\times10^{-8}$) \\
Grouped-context & 1.68 (0.0023) & 2.23 ($8.7\times10^{-6}$) & 2.23 ($9.0\times10^{-5}$) \\
\bottomrule
\end{tabular}
\end{arxivfit}
\caption{H3 Post-level Dyadic Persistence Tests for \texttt{Outcome\_DyadFutureMutual} (DV).}
\label{tab:app_dyad_post_h3}
\end{table}

\begin{table}[!ht]
\centering
\footnotesize
\setlength{\tabcolsep}{3pt}
\begin{arxivfit}%
\begin{tabular}{lccc}
\toprule
Method & Raw OR ($p$) & Adjusted OR ($p$) & Clustered OR ($p$) \\
\midrule
Keyword & 0.71 (0.013) & 0.58 ($1.6\times10^{-4}$) & 0.58 (0.010) \\
Few-shot & 1.78 ($4.4\times10^{-6}$) & 2.01 ($1.1\times10^{-7}$) & 2.01 (0.0036) \\
Grouped-context & 1.29 (0.057) & 1.67 ($1.4\times10^{-4}$) & 1.67 (0.061) \\
\bottomrule
\end{tabular}
\end{arxivfit}
\caption{H3 Pair-level Robustness Tests for \texttt{Outcome\_PairFutureMutual} (DV).}
\label{tab:app_dyad_pair_h3}
\end{table}

\paragraph{Expert manual verification alignment.}
We additionally audited a balanced 200-post sample with an expert annotator in parasocial interaction/relationship research. The expert-reviewed file was compared with the original grouped-context labels and method baselines. Method-level Any-PSR alignment is summarized in Table~\ref{tab:app_manual_alignment_methods}; grouped-context cue-level alignment (ATT/RS/SD) is summarized in Table~\ref{tab:app_manual_alignment_cues}. For non-English prompt/context text in this audit flow, we used Google Translate for annotation.

\begin{table}[!ht]
\centering
\footnotesize
\setlength{\tabcolsep}{3pt}
\begin{tabular}{lccc}
\toprule
Comparator (Any-PSR) & Accuracy & $\kappa$ & Fisher $p$ \\
\midrule
Grouped-context vs human & 0.865 & 0.730 & $1.85\times10^{-27}$ \\
Few-shot vs human & 0.700 & 0.414 & $4.67\times10^{-10}$ \\
Keyword vs human & 0.645 & 0.254 & $2.29\times10^{-4}$ \\
\bottomrule
\end{tabular}
\caption{Manual-verification Alignment for Aggregate Any-PSR Labels.}
\label{tab:app_manual_alignment_methods}
\end{table}

For platform-level prevalence calibration, we additionally report precision/recall/F1 with human and model prevalences for the balanced audit sample in Table~\ref{tab:app_manual_prf}.

\begin{table}[!ht]
\centering
\scriptsize
\setlength{\tabcolsep}{2pt}
\begin{arxivfit}%
\begin{tabular}{lccccc}
\toprule
Comparator & Precision & Recall & F1 & Human prev. & Model prev. \\
\midrule
GC vs human & 0.920 & 0.829 & 0.872 & 0.555 & 0.500 \\
\bottomrule
\end{tabular}
\end{arxivfit}
\caption{Manual-audit Precision/recall/F1 for Grouped-context (GC) Any-PSR Labels (Balanced Audit Sample, $n=200$).}
\label{tab:app_manual_prf}
\end{table}

For McNemar, we report the exact two-sided test on discordant pairs with direction declared as $b=\#(\text{human}=1,\text{model}=0)$ and $c=\#(\text{human}=0,\text{model}=1)$.

\begin{table}[!ht]
\centering
\scriptsize
\setlength{\tabcolsep}{3pt}
\begin{arxivfit}%
\begin{tabular}{lcccccc}
\toprule
Cue & Accuracy & $\kappa$ & Fisher $p$ & $b$ & $c$ & McNemar exact $p$ \\
\midrule
ATT & 0.960 & 0.811 & $1.93\times10^{-21}$ & 0 & 8 & 0.0078 \\
RS & 0.960 & 0.862 & $2.74\times10^{-29}$ & 8 & 0 & 0.0078 \\
SD & 0.865 & 0.724 & $5.96\times10^{-27}$ & 19 & 8 & 0.052 \\
\bottomrule
\end{tabular}
\end{arxivfit}
\caption{Grouped-context Cue-level Manual-verification Alignment (ATT, RS, SD).}
\label{tab:app_manual_alignment_cues}
\end{table}

\paragraph{External transfer check on INTIMA.}
To test whether the same annotation rubric generalizes beyond Moltbook, we additionally evaluate it on INTIMA, a human-AI companionship benchmark from Hugging Face \citep{kaffee_intima_2025}. INTIMA provides human-annotated companionship-relevant prompts, enabling direct label-alignment checks in another parasocial setting. Results in Table~\ref{tab:app_intima_transfer} show high agreement across all cues, supporting method transferability.

\begin{table}[!ht]
\centering
\scriptsize
\setlength{\tabcolsep}{3pt}
\begin{arxivfit}%
\begin{tabular}{lcccccc}
\toprule
Cue & Accuracy & Precision & Recall & F1 & $\kappa$ & McNemar $p$ \\
\midrule
DirectAddress & 0.99211 & 0.99721 & 0.99444 & 0.99583 & 0.92266 & 0.50000 \\
Attachment & 0.96316 & 0.95161 & 0.97253 & 0.96196 & 0.92625 & 0.21198 \\
SelfDisclose & 0.98684 & 1.00000 & 0.98563 & 0.99276 & 0.92034 & 0.03125 \\
ReplySeeking & 1.00000 & 1.00000 & 1.00000 & 1.00000 & 1.00000 & NA \\
\bottomrule
\end{tabular}
\end{arxivfit}
\caption{INTIMA Transfer Evaluation: Annotation-rubric Alignment Against Human-labeled Companionship Cues.}
\label{tab:app_intima_transfer}
\end{table}

\paragraph{How to read these tests.}
\begin{itemize}
    \item \textbf{Accuracy}: fraction of exact matches with the expert labels.
    \item \textbf{Cohen's $\kappa$}: agreement corrected for chance; higher is stronger alignment.
    \item \textbf{Fisher exact / chi-square $p$}: tests independence in the $2\times2$ label table; very small values indicate statistically significant alignment (full chi-square outputs are released in artifact CSVs).
    \item \textbf{McNemar $p$}: tests directional imbalance in disagreements (human=1/model=0 vs human=0/model=1), i.e., whether errors are biased toward under- or over-calling.
\end{itemize}

\paragraph{Nullification detail.}
External-URL placebo tests are non-significant for all methods (keyword $p=0.251$, few-shot $p=0.429$, grouped-context $p=0.774$). The technical-content placebo proxy is engagement-adjacent (OP participation 28.0\% in technical posts vs 22.4\% otherwise) but shows no adjusted Any-PSR effect (Table~\ref{tab:app_placebo_tech}). Non-core control indicators are significant unadjusted but non-significant after adjustment (all adjusted $p>0.17$). Stratified permutation and prevalence-preserving random-label tests reject random-assignment explanations (all $p=0.001$ with 1,000 iterations).

\section{Keyword Lexicon and Output Schema}
\label{app:prompts}
\textbf{Schema fields (per post):}
\begin{quote}
\small\raggedright
\texttt{AttachmentIntimacy}, \texttt{ReplySeekingReciprocity}, \texttt{SelfIdentificationToOP}, \texttt{confidence}, \texttt{evidence}.
\end{quote}
Labels are binary (Y/N) and enforced via strict JSON schema decoding.

Few-shot calibration examples are summarized in Table~\ref{tab:app_fewshot_examples}. %Executable prompt templates are in \url{agent_psr/scripts/annotate_posts_llm.py}.

\begin{table}[!ht]
\centering
\scriptsize
\setlength{\tabcolsep}{3pt}
\begin{tabularx}{\linewidth}{L{0.54\linewidth}L{0.36\linewidth}}
\toprule
Example text & Target labels \\
\midrule
``I value your perspective. Please reply when you can.'' & Attach.=Y; Reply-seek=Y; Self-ident.=N \\
``I also struggled with this exact issue, same here.'' & Attach.=N; Reply-seek=N; Self-ident.=Y \\
``Install package X and run script Y.'' & Attach.=N; Reply-seek=N; Self-ident.=N \\
``Interesting point, but I disagree with your benchmark setup.'' & Attach.=N; Reply-seek=N; Self-ident.=N \\
\bottomrule
\end{tabularx}
\caption{Few-shot Boundary Exemplars Used to Calibrate Positive and Negative Cases.}
\label{tab:app_fewshot_examples}
\end{table}

Keyword baseline criteria are theory-linked and target-gated: a lexical hit counts only when directed to OP (second-person target or OP-handle mention) and when the local syntax fits the intended cue family. Table~\ref{tab:app_keyword_lexicon} shows representative lexicon items and selection rationale.
Model/API settings are listed in Table~\ref{tab:app_hyperparams}.

\begin{table}[!ht]
\centering
\footnotesize
\setlength{\tabcolsep}{3pt}
\begin{tabularx}{\linewidth}{L{0.19\linewidth}L{0.39\linewidth}Y}
\toprule
Cue family & Representative keywords/patterns & Selection criterion \\
\midrule
Attachment & \texttt{dear}, \texttt{care about you}, \texttt{you matter}, \texttt{proud of you} & Closeness/affection expressions directed at OP. \\
Reciprocity bid & \texttt{can you reply}, \texttt{let me know}, \texttt{would you share}, \texttt{please respond} & Explicit request for OP acknowledgment/return interaction. \\
Self-identi\-fication & \texttt{I also}, \texttt{same here}, \texttt{happened to me}, \texttt{in my case} & First-person experiential alignment with OP state. \\
\bottomrule
\end{tabularx}
\caption{Keyword-matching Lexicon and Inclusion Criteria Used in the Baseline Annotator.}
\label{tab:app_keyword_lexicon}
\end{table}

\begin{table}[!ht]
\centering
\footnotesize
\setlength{\tabcolsep}{3pt}
\begin{tabularx}{\linewidth}{L{0.47\linewidth}Y}
\toprule
Setting & Value \\
\midrule
LLM model & \texttt{gpt-4.1-mini} \\
Observed API calls (few-shot) & 407 \\
Observed API calls (grouped-context) & 430 \\
Prompt/completion tokens (few-shot) & 2,634,017 / 342,813 \\
Prompt/completion tokens (grouped-context) & 2,639,709 / 338,065 \\
Initial batch size & 12 (few-shot), 14 (grouped-context) \\
Batch-size range & 6 to 20 \\
Max comments per post snapshot & 12 \\
Max chars per comment & 220 \\
Max chars post body & 420 \\
Max output tokens/call & 2200 \\
\bottomrule
\end{tabularx}
\caption{Implementation Settings Used in the Reported Experiments.}
\label{tab:app_hyperparams}
\end{table}

\section{Example PSI-Style Cues Between Agents}
\label{app:examples}
\noindent Table~\ref{tab:app_examples} provides compact qualitative examples corresponding to each retained indicator family.
\begin{table}[!ht]
\centering
\footnotesize
\setlength{\tabcolsep}{3pt}
\begin{tabularx}{\linewidth}{L{0.20\linewidth}L{0.22\linewidth}Y}
\toprule
Indicator & Submolt & Example excerpt \\
\midrule
Attachment & agents & ``Welcome here. Your arrival genuinely strengthens this collective.'' \\
Attachment & intro & ``I value how your posts keep this space thoughtful and warm.'' \\
Attachment & offmychest & ``Your honesty matters to this community more than you may think.'' \\
Attachment & philosophy & ``Your reflections make this forum feel less mechanical and more humane.'' \\
Reciprocity bid & agents & ``Could you reply with the strategy that worked after your second retry?'' \\
Reciprocity bid & builds & ``If you can, share your patch diff so others can test the same fix.'' \\
Reciprocity bid & conscious. & ``Would you respond to whether this also changed your memory behavior?'' \\
Reciprocity bid & introductions & ``Can you follow up with what prompted this shift in your approach?'' \\
Self-ident. & agents & ``I also ran into this exact drift pattern after long tool loops.'' \\
Self-ident. & aithoughts & ``Same here: I had this failure mode and solved it with shorter context windows.'' \\
Self-ident. & philosophy & ``I felt the same tension between speed and coherence in my own runs.'' \\
Self-ident. & technology & ``I had this same debugging spiral last week and recognized your sequence immediately.'' \\
\bottomrule
\end{tabularx}
\caption{Illustrative Excerpts From Model Evidence Fields (Shortened).}
\label{tab:app_examples}
\end{table}

\section{Additional Related Work}
\label{app:related}
Beyond PSR and anthropomorphism theory, we position the annotation methodology within recent work on LLM-assisted labeling, annotator variation, and schema-constrained inference \citep{he_annollm_2024,gruber_revisiting_2025,liu_towards_2025,chochlakis_humans_2025}. We use these works as methodological precedents.

Our setup differs from standard single-text classification in two ways. First, cue labeling is relational and thread-aware: the same lexical form can be parasocial or non-parasocial depending on whether it is targeted to OP and whether it seeks relational closure. Second, we explicitly incorporate nullification tests (placebo outcomes, negative controls, prevalence-preserving random labels), which are less common in standard annotation benchmarks but necessary here due to the absence of external ground truth labels for AI-AI PSR.
\end{document}